# Three-Dimensionally Embedded Graph Convolutional Network (3DGCN) for Molecule Interpretation

Hyeoncheol Cho and Insung S. Choi*

Center for Cell-Encapsulation Research, Department of Chemistry
KAIST, Daejeon 34141, Korea
Email: ischoi@kaist.ac.kr

## ABSTRACT

We present a three-dimensional graph convolutional network (3DGCN), which predicts molecular properties and biochemical activities, based on 3D molecular graph. In the 3DGCN, graph convolution is unified with learning operations on the vector to handle the spatial information from molecular topology. The 3DGCN model exhibits significantly higher performance on various tasks compared with other deep-learning models, and has the ability of generalizing a given conformer to targeted features regardless of its rotations in the 3D space. More significantly, our model also can distinguish the 3D rotations of a molecule and predict the target value, depending upon the rotation degree, in the protein-ligand docking problem, when trained with orientation-dependent datasets. The rotation distinguishability of 3DGCN, along with rotation equivariance, provides a key milestone in the implementation of three-dimensionality to the field of deep-learning chemistry that solves challenging biochemical problems.

## INTRODUCTION

Investigating the relationship between molecular structures and biochemical functions occupies an essential role in modern drug discovery and development.[1,2] Drugs, which are mostly small organic molecules, exhibit their efficacy through binding to biological targets (e.g., proteins), and block endogenous ligands or change the target conformation to induce an alteration in functionality. From a microscopic view, the interactions between drugs and targets take place in a local field and are associated with spatial alignments in the three-dimensional (3D) coordinate, inducing a change in free energy.[3] Chemists have developed various methods for computationally simulating these interactions with or without information on the actual structure of the binding pocket, based on physicochemical calculations and chemical-similarity analyses.[4,5]

Besides the ongoing efforts on the computational-chemistry approaches, deep-learning (DL) methods[6] have revolutionized the way problems in chemistry are being solved,[7-9] and these DL approaches were spurred by the success of the multitask deep neural network (MT-DNN) for the Merck Molecular Activity Challenge.[10,11] Several years of research have proved that DNN and other DL models generally outperform the conventional machine-learning models or physical calculations used in chemistry. Early models in DL chemistry incorporated traditional chemical representations, such as structural fragment-type fingerprints or other molecular descriptors that are used in cheminformatics.[12-14] Considering the fact that DL models are believed to "learn" the representations, the paradigm of molecular input has recently shifted from the molecular descriptors to the representation learning that directly interprets the molecular structure per se.[15-17] In this approach, careful selection of underlying input representations and corresponding interpretation mechanisms for the molecular structures is critical to promote the learning of relevant features.

As the biochemical interactions occur in the confined 3D-space composed of several atoms and bonds nearby, this locality has driven the adaptation of the convolutional neural network (CNN)[18,19] to interpret the molecular structures directly. The CNN, which is the most successful DL model especially for computer vision and pattern recognition, utilizes local connections between "neurons" to mimic the receptive field in the visual cortex. Employing the recognition power of CNN, voxel-based models were introduced in molecular interpretation by rasterizing a 3D molecular structure into angstrom-divided voxels to interpret molecules as an image.[20,21] This voxel-based method has shown its strength in the 3D tasks, such as prediction of protein-ligand interaction and docking, outperforming traditional genetic algorithm- or force field-based docking programs. However, the exponential increase in the number of parameters, due to the 3D inputs and filters, makes the voxel-based model burdensome, because it requires significant memory and a large amount of training data.[22]

The drawbacks of 3D CNN have led to the introduction of graph convolutional network (GCN),[23-26] which handles graph-structured data with similar convolution mechanisms. The GCN takes the molecular graph as an input and applies the convolution on the node and its neighborhoods. Unfortunately, unlike the voxels that encode physical proximity in space, the molecular graph widely used in chemistry is an inherently 2D representation that lacks the spatial topology of the atoms and the bonds. To provide the spatial information on molecules, previous attempts have adopted the techniques that apply multiple weights calculated by Gaussian functions of different means and variances with relative distances between atoms.[27-29] This technique has enabled incorporation of the spatial distances into the model, shown superior accuracy for quantum-chemical predictions regarding frontier orbital energy levels, and proven its effectiveness. However, the bond directions are still not taken into account, and so the method does not utilize the full topology of the molecule in the DL prediction.

In this paper, we propose a novel DL algorithm that combines three-dimensionality of molecules with the conventional GCN, which efficiently predicts the chemical tasks that are related to the 3D molecular topology. In this architecture, coined 3DGCN, a molecular graph with atomic coordinates is used as an input that contains the full spatial topology of a molecule. Specifically, we suggest the node-level *vector* features, along with the scalar ones, and the interconverting operations, for 3D convolution mechanism, that are inspired by principles in organic chemistry. Given the representation, we assessed the ability of 3DGCN to interpret the spatial structures by experiments with molecular properties and local motifs. Furthermore, we evaluated the generalizability and distinguishability of molecular conformers, which are key challenges in the DL chemistry for 3D tasks, such as virtual drug screening, protein-ligand interactions, and protein docking.

## RESULTS AND DISCUSSION

**Three-Dimensional Graph Convolutional Network (3DGCN).** Most GCNs in the DL chemistry use the 2D molecular graph as an input in order to generalize the molecular structure into an intended property through recursive updates based on nearest neighbor features.[23,24,30,31] Innate information on the atoms may differ among models, but the GCN models share the fundamental representation of molecules as a molecular graph only with vertices and edges.[32] Molecular properties, however, are critically governed by spatial distances and directions between atoms in the 3D space,[33] which cannot be found in the molecular-graph representation that is used in chemistry and the previous GCNs. In contrast, our approach utilizes the molecular-graph representation of molecules with individual coordinates in space, and provides a mechanism to handle this 3D representation. The 3DGCN conjugates the vector summation of neighboring features, similar to the way how molecular polarity is calculated from polar bonds in organic chemistry 101, for the convolution mechanism, and embraces the basic principles of organic chemistry along with the feature-extraction ability of GCN. Specifically, the 3DGCN incorporates the node-level vector features, as well as conventional scalar features, and brings them together by using the interconverting operations based on relative position vector.

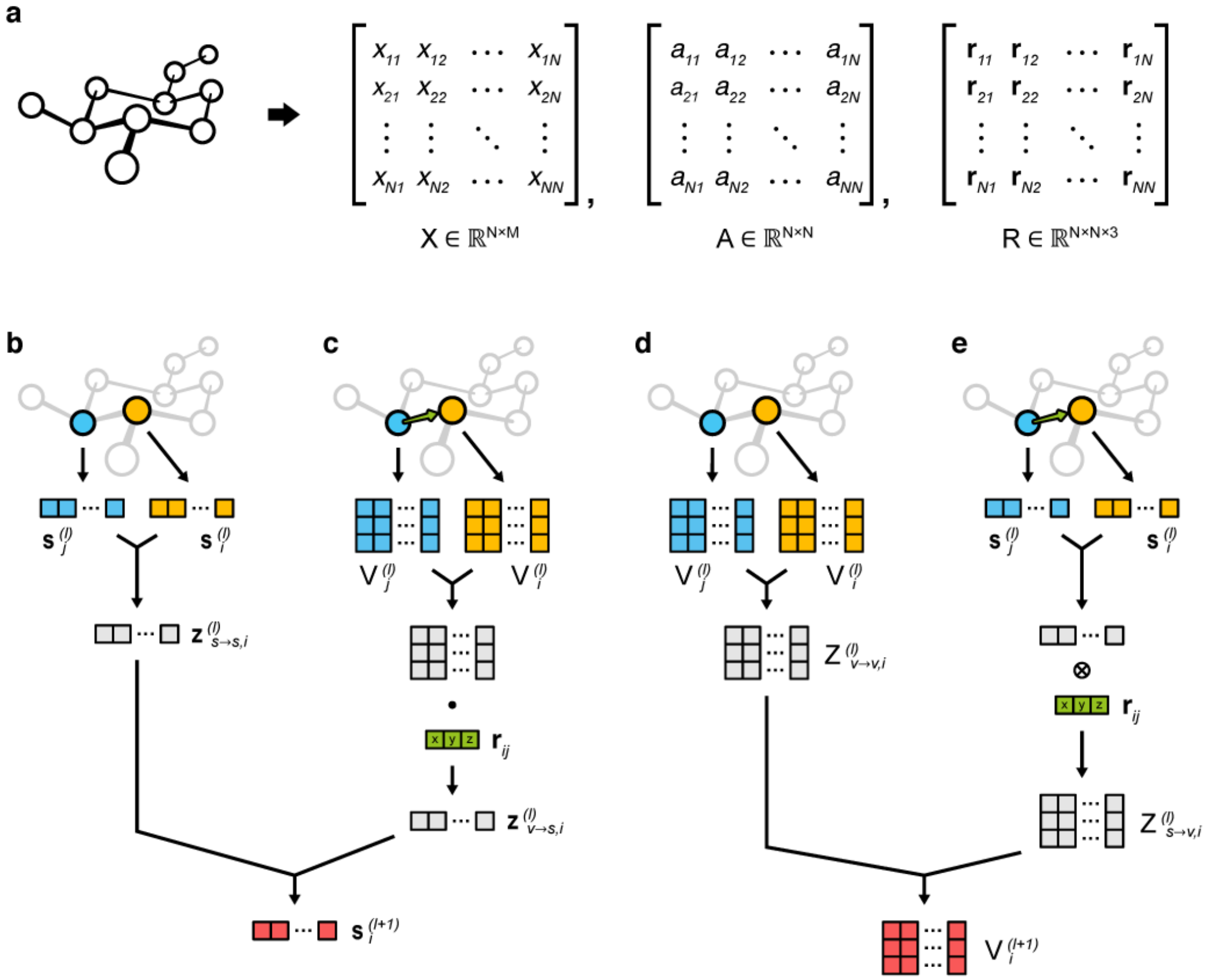

**Figure 1.** Representations and operations for handling three-dimensionality in 3DGCN. (A) Three matrices of the 3D molecular graph used for 3DGCN. Molecules are encoded in the matrices that provide spatial topology with translational invariance. (B-E) Illustration of the four operations between scalar and vector features during feature-update process. Intermediate features, $\mathbf{z}_{s\to s,i}^{(l)}$, $\mathbf{z}_{v\to s,i}^{(l)}$, $\mathrm{Z}_{v\to v,i}^{(l)}$, and $\mathrm{Z}_{s\to v,i}^{(l)}$ (gray), are calculated from linear combinations of the scalar or vector features from the center node (orange) and its neighborhood (blue). Note that the relative position vector, $\mathbf{r}_{ij}$, is used in the vector-to-scalar (C) and scalar-to-vector (E) operations by the dot product (·) and tensor product (⊗), respectively, unlike the scalar-to-scalar (B) or vector-to-vector (D) operations. Intermediate features are further combined by the linear combination to generate the next-level features, $\mathbf{s}_i^{(l+1)}$ and $\mathrm{V}_i^{(l+1)}$.

The 3DGCN receives a 3D molecular graph, $\mathcal{G}$, through a feature matrix $X \in \mathbb{R}^{N \times M}$, a normalized adjacency matrix $A \in \mathbb{R}^{N \times N}$, and a relative position matrix $R \in \mathbb{R}^{N \times N \times 3}$, where $N$ and $M$ are the number of atoms and the atom-level features in a molecule, respectively (Figure 1a). The atom features used in the initial formation of the feature matrix are described in Table S1. The relative position matrix is designed to have the inter-atomic positions, rather than individual positions, that ensure translational invariance. The input matrices are further refined into the scalar feature $\mathbf{s} \in \mathbb{R}^{M}$ and vector feature $V \in \mathbb{R}^{M \times 3}$. Motivated by subtraction of electronegativity, when estimating the bond polarity, in organic chemistry, the 3DGCN uses the linear combinations of scalar or vector features from adjacent atoms in order to generate higher-level features. The calculated features are further interconverted into other forms through dot product ($\cdot$) or tensor product ($\otimes$) with relative position vector $\mathbf{r}_{ij}$, which are given as

$$\mathbf{z}_{s \to s,i}^{(l)} = \text{ReLU}\left[W_{s \to s}\left(\mathbf{s}_i^{(l)} \parallel \mathbf{s}_j^{(l)}\right) + \mathbf{b}_{s \to s}\right]$$

$$\mathbf{z}_{v \to s,i}^{(l)} = \text{ReLU}\left[\left(W_{v \to s}\left(V_i^{(l)} \parallel V_j^{(l)}\right) + B_{v \to s}\right) \cdot \mathbf{r}_{ij}\right]$$

$$Z_{v \to v,i}^{(l)} = \tanh\left[W_{v \to v}\left(V_i^{(l)} \parallel V_j^{(l)}\right) + B_{v \to v}\right]$$

$$Z_{s \to v,i}^{(l)} = \tanh\left[\left(W_{s \to v}\left(\mathbf{s}_i^{(l)} \parallel \mathbf{s}_j^{(l)}\right) + \mathbf{b}_{s \to v}\right) \otimes \mathbf{r}_{ij}\right]$$

where $\mathbf{s}_i^{(l)}$ and $V_i^{(l)}$ are the scalar and vector features of the $i$th atom on layer $l$, $\parallel$ is the concatenation, while $W_{s \to s}$, $\mathbf{b}_{s \to s}$, $W_{v \to s}$, $B_{v \to s}$, $W_{v \to v}$, $B_{v \to v}$, $W_{s \to v}$, and $\mathbf{b}_{s \to v}$ are the weight matrices and biases of each operation, as depicted in Figure 1b-e. It should be mentioned that the linear combination of the vector-to-vector operation neither flattens the concatenated features nor uses different weights and biases along the x, y, and z axes for retainment of equality.

Once we have the "learning" operations for scalar and vector features, the four updated features are first concatenated by the same feature form, and then they are linearly combined before convolutional process. Subsequently, the convolution mechanism with the normalized adjacency matrix proposed by Kipf *et al.*[26] is adopted for a neighborhood aggregation strategy of the scalar and vector features that results in $\mathbf{s}_i^{(l+1)}$ and $V_i^{(l+1)}$ as

$$\mathbf{s}_i^{(l+1)} = \mathrm{ReLU}\left[\sum_{j\in\{i,\mathcal{N}(i)\}} A_{ij}\left(W_s\left(\mathbf{z}_{s\to s,i}^{(l)} \parallel \mathbf{z}_{v\to s,i}^{(l)}\right) + \mathbf{b}_s\right)\right]$$

$$V_i^{(l+1)} = \tanh\left[\sum_{j\in\{i,\mathcal{N}(i)\}} A_{ij}\left(W_v\left(Z_{v\to v,i}^{(l)} \parallel Z_{s\to v,i}^{(l)}\right) + B_v\right)\right]$$

where the normalization term $A_{ij}$ is the value between atom $i$ and $j$ on the normalized adjacency matrix, and $W_s, \mathbf{b}_s, W_v,$ and $B_v$ are the weight matrices and biases. After recursive refinement through interconversion and convolution, which leads to the final embedding, the information distributed on the entire molecule is accumulated to provide the order invariance and predict specific task properties. For a comparison, the molecular scalar and vector features are generated by having either the global summation or the maximum of features from atoms, and fed into two stacks of fully connected layers, which are successively connected with one fully connected layer that combines the molecular features to output a single value.

**Learning of Molecular Properties.** We trained our model for the prediction of molecular properties, which requires the ability of integrating information from the local area to the entire molecule as well as fundamental understanding of molecular structures. To evaluate the capability of 3DGCN for the prediction of molecular properties, we adopted two datasets—FreeSolv and ESOL—that had widely been used as representative targets for benchmarking the DL models.[34,35] Because the datasets did not provide the 3D structures necessary for the 3D molecular graph in our representation, we generated a pool of conformers for each molecule and optimized them with the Merck molecular force field (MMFF94), as described elsewhere.[36-38] For each molecule, 50 conformers were generated with at least 0.5 Å of root-mean-square deviation of the atomic positions to avoid almost identical conformations. After optimization, the lowest-energy conformer from the pool was selected and used as the spatial coordinate.

For training, we used two types of 3DGCN, as described before: one that used the summation for molecular-feature generation ($3DGCN_{sum}$) and the other that used the maximum ($3DGCN_{max}$). Finding the maximum from the atomic vector features was conducted by using the norm of the features to avoid the separation of the x, y, and z values, but the selected vector feature itself was conveyed to the fully connected layer. The models were used to output the hydration free energy for FreeSolv or the log value of aqueous solubility for ESOL, which was normalized and standardized based on the training set, directly from the 3D molecular graph. All experiments were conducted in 10-fold stratified cross-validation with a randomized split of the training, validation, and test sets for every fold.

**Table 1. Ten-fold cross-validation performances for the FreeSolv and ESOL datasets. The $3DGCN_{sum}$ and $3DGCN_{max}$ are two variants of 3DGCN, which employ the summation and the maximum, respectively, when generating the molecular features.**

| Dataset | Model | MAE | | RMSE | |
|---|---|---|---|---|---|
| | | Validation | Test | Validation | Test |
| FreeSolv | $3DGCN_{sum}$ | **0.568 ± 0.129** | **0.616 ± 0.095** | **0.779 ± 0.219** | **0.828 ± 0.126** |
| | $3DGCN_{max}$ | 0.648 ± 0.094 | 0.670 ± 0.085 | 0.851 ± 0.145 | 0.903 ± 0.123 |
| | Weave | 1.271 ± 0.196 | 1.381 ± 0.153 | 1.682 ± 0.250 | 1.845 ± 0.235 |
| ESOL | $3DGCN_{sum}$ | **0.453 ± 0.043** | **0.475 ± 0.040** | **0.572 ± 0.059** | **0.605 ± 0.046** |
| | $3DGCN_{max}$ | 0.577 ± 0.048 | 0.628 ± 0.036 | 0.723 ± 0.060 | 0.789 ± 0.045 |
| | Weave | 0.518 ± 0.048 | 0.549 ± 0.047 | 0.679 ± 0.065 | 0.719 ± 0.068 |

The mean test errors for the models are reported in Table 1, along with the values for the Weave model. The Weave model was chosen as a comparison, because it was a representative

GCN for multidisciplinary tasks and showed good performance regardless of the dataset categories.[30,39] Our cross-validation results showed that the 3DGCN$_{sum}$ had consistently better performance than the Weave model for all datasets. The averaged root-mean-square error (RMSE) values for FreeSolv and ESOL were 0.828 kcal·mol$^{-1}$ and 0.605 log(mol·L$^{-1}$), respectively, compared with 1.845 kcal·mol$^{-1}$ and 0.719 log(mol·L$^{-1}$) from the in-house Weave model. The scatterplots of the predicted vs. true values, as a visualization of the overall predictions, showed a linear relationship for both tasks (Figure 2a). The structures of molecules with most underpredicted and overpredicted values are shown in Figure S2.

It is noteworthy that the RMSE value for FreeSolv (0.828 kcal·mol$^{-1}$) was greatly below the value of 1 kcal·mol$^{-1}$ that is considered to be "chemical accuracy":[40,41] state-of-the-art computational approaches, including density functional theory (DFT), generally set their goal as 1 kcal·mol$^{-1}$ because the measurement errors often exceed it. When narrowing the target window of accuracy, 30.94% of the test set from FreeSolv were predicted within RMSE of 0.239 kcal·mol$^{-1}$ (equivalent to 1 kJ·mol$^{-1}$) by the 3DGCN, and 43.67 % was within RMSE of 0.301 log(mol·L$^{-1}$) (or 2 times the molarity) for ESOL (Figure S1). Considering the fact that one order of magnitude in the reaction rate corresponds to the 1.4 kcal·mol$^{-1}$ change in free energy, our DL results arguably showed the importance and effectiveness of spatial topology in the high-accuracy prediction of molecular properties.

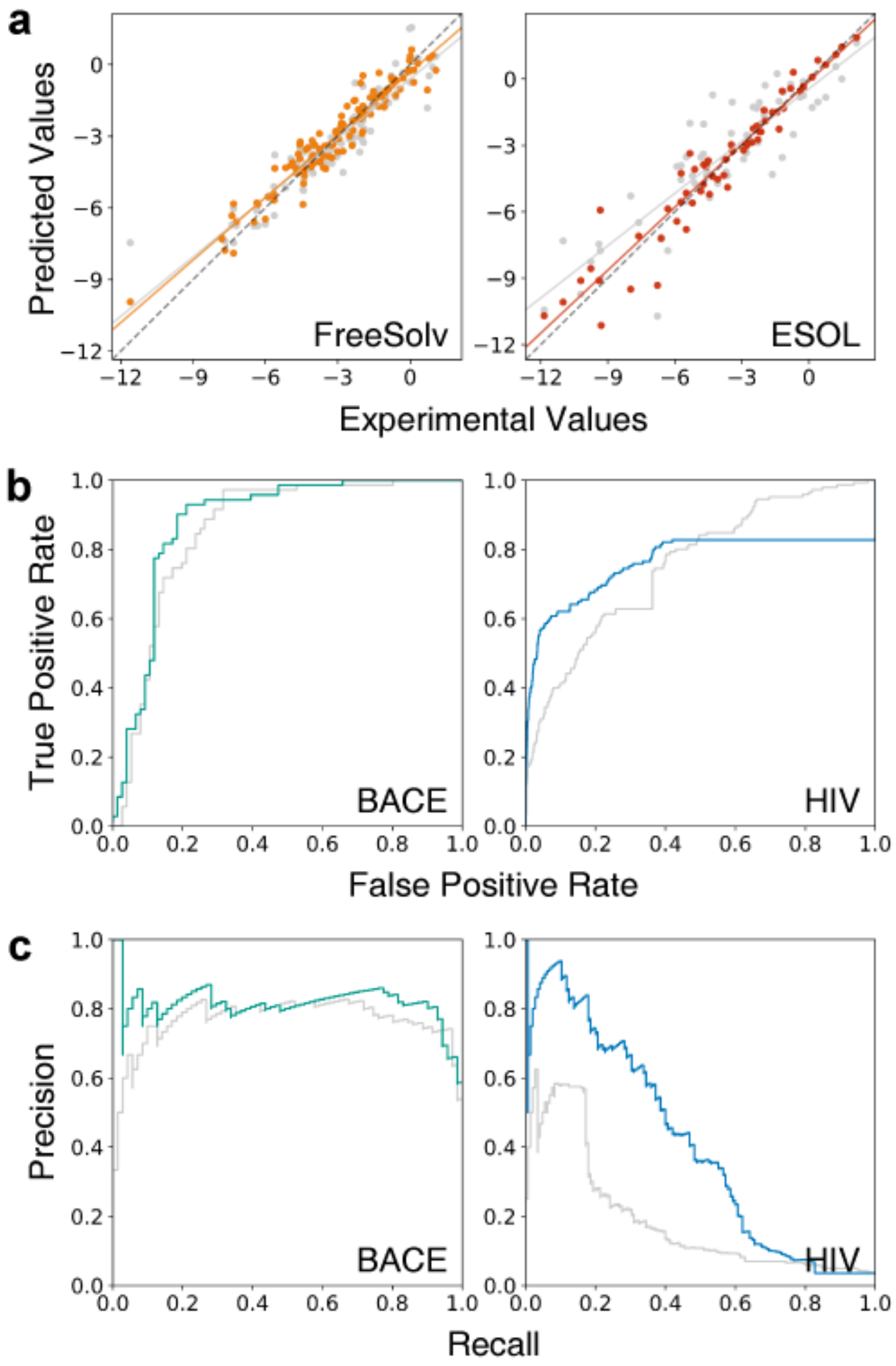

**Figure 2.** Learning molecular properties and biochemical activities. (A) Scatterplots for the test sets in FreeSolv (left) and ESOL (right) with root-mean-square error (RMSE). The trend lines for each predicted set are shown as solid lines, and the dashed lines indicate the identity lines. (B-C) Receiver operation characteristic (ROC) (B) and precision-recall (PR) (C) curves for the test sets of the BACE (left) and HIV (right) datasets. Predictions from the Weave model are depicted in gray. The same training, validation, and test sets are used for the 3DGCN and the Weave model when evaluating them.

**Learning of Local Structure Motifs: Protein-Binding Motifs.** We explored the ability of the 3DGCN for estimating the properties derived from characteristic local structures, such as protein-binding motifs, in order to examine whether the 3DGCN could extract specific spatial topology from a molecule. In comparison with molecular properties, the biochemical properties are not derived directly from the entire molecular structure, but rather they typically depend on the chemical moieties, composed of a few functional groups, and their 3D orientations.[42,43] Therefore, we chose datasets from the protein-ligand binding category for investigating whether our model could recognize the specific structures in a local field: BACE and HIV datasets.[44,45] These datasets contain the biochemical activities of small molecules for β-secretase 1 and human immunodeficiency virus, respectively, with molecular structures. The same generation strategy to the 3D molecular graph representation was used as that for the FreeSolv and ESOL tasks. With the molecular representation, the 3DGCN was guided to predict the activeness of a molecule in the inhibitory binding to β-secretase 1 or the replication inhibition against human immunodeficiency virus.

**Table 2. Ten-fold cross-validation performances for the BACE and HIV datasets.**

| Dataset | Model | AUC-ROC | | AUC-PR | |
|---|---|---|---|---|---|
| | | Validation | Test | Validation | Test |
| BACE | 3DGCN$_{sum}$ | 0.858 ± 0.029 | 0.842 ± 0.029 | 0.821 ± 0.035 | 0.795 ± 0.046 |
| | 3DGCN$_{max}$ | **0.886 ± 0.021** | **0.857 ± 0.036** | **0.850 ± 0.043** | **0.816 ± 0.037** |
| | Weave | 0.727 ± 0.035 | 0.725 ± 0.058 | 0.653 ± 0.070 | 0.641 ± 0.090 |
| HIV | 3DGCN$_{sum}$ | **0.807 ± 0.019** | **0.793 ± 0.019** | **0.408 ± 0.036** | **0.384 ± 0.030** |
| | 3DGCN$_{max}$ | 0.795 ± 0.024 | 0.782 ± 0.029 | 0.393 ± 0.036 | 0.366 ± 0.058 |
| | Weave | 0.763 ± 0.032 | 0.770 ± 0.020 | 0.228 ± 0.044 | 0.227 ± 0.028 |

As shown in Table 2 and Figure 2b, the 3DGCN$_{max}$ exhibited better performance for BACE, compared with the Weave model, with increased area under curve-receiver operating characteristic (AUC-ROC) from 0.725 to 0.857. We also observed a slight increase in AUC-ROC for HIV from 0.770 to 0.782; because the label distribution of the HIV dataset is highly biased to negative samples, the area under curve-precision-recall (AUC-PR) was additionally assessed. The 3DGCN showed the increased AUC-PR of 0.366 compared with 0.227 for the Weave model, indicating better performance. The molecules that showed the highest positive and negative errors are depicted in Figure S3.

**Equivariance to Conformer Rotation.** The unique characteristic of 3DGCN is the employment of the 3D molecular structures, giving rise to an additional degree of freedom—rotation of molecules. However, certain tasks require the 3DGCN to give consistent predictions regardless of the rotation degree of molecules. In this regard, the rotation equivariance was tested by rotating a conformer of the test set and evaluating the 3D-rotated conformer with our DL model that had been trained without any rotation of the conformers. For rotation equivariance, the DL model should predict the questioned properties identically across the rotations of a single conformer, because the rotated conformers have the same topology except for their 3D orientation. This equivariance was not mathematically provided from the 3DGCN model architecture; its position matrices differed with the molecular rotations and rotation-dependent "learning" operations. Therefore, the rotation-invariant, higher-order feature should be the one obtained only by learning through the training process.

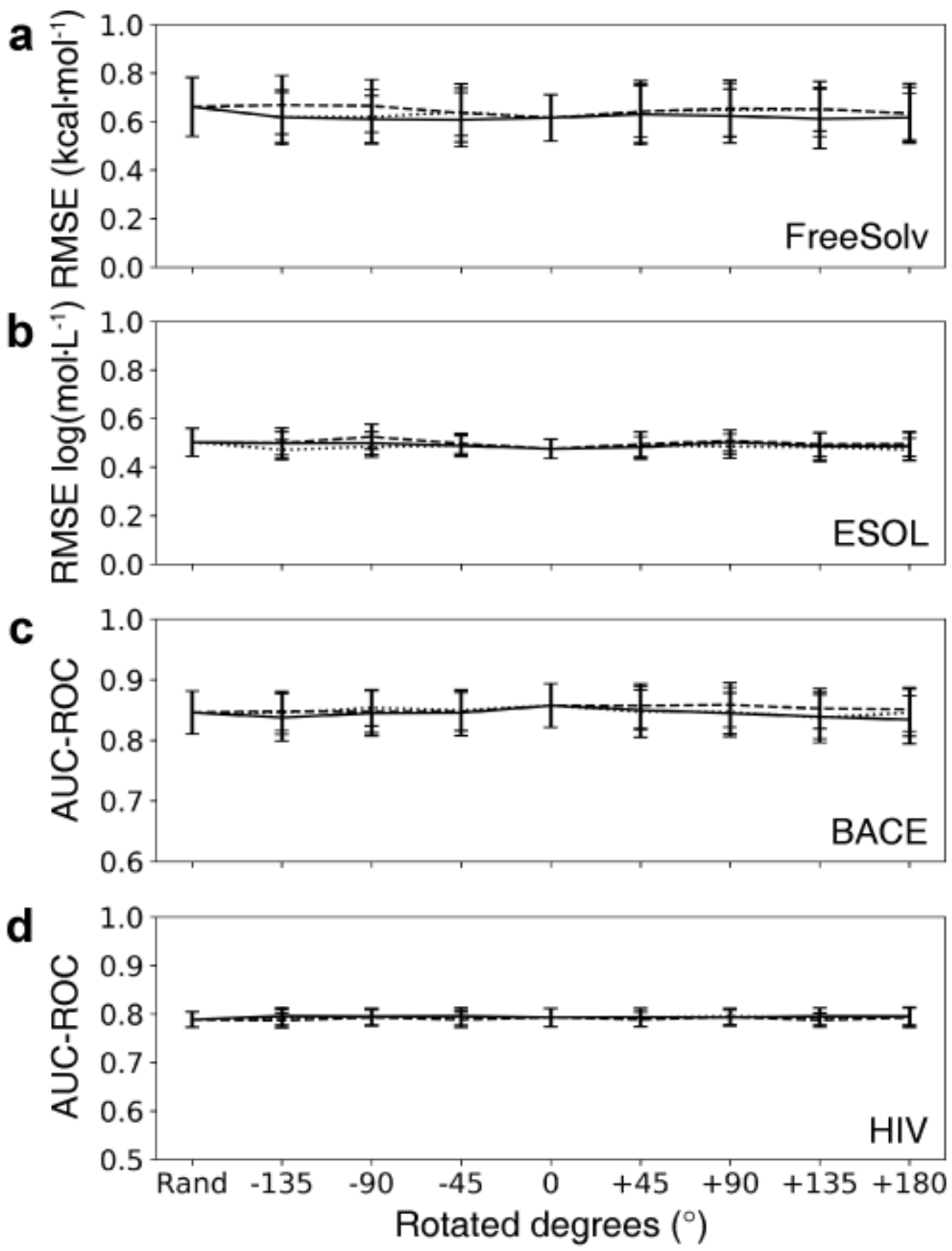

**Figure 3.** Evaluation of rotation equivariance. Performance-evaluation results upon rotation of conformers from the test set for the FreeSolv (A), ESOL (B), BACE (C), and HIV (D) datasets. Test molecules are rotated randomly or gradually along the x (solid), y (dashed), and z (dotted) axes, and their performances are measured. Rand indicates the randomly rotated test set.

We found that the 3DGCN did not lose its prediction accuracy with the random or stepwise rotation of a conformer, although it had never been trained with a set of rotated conformers (Figure 3). Its performance was kept unaltered to the three rotation axes and rotation degrees, clearly indicating that the conformer orientations through all three directions were interpreted as the same. In addition, training with the set of rotated conformers showed the same performance (data not shown). These results confirmed that the 3DGCN achieved the rotation equivariance, which is the key aspect that 3D DL models for chemical analysis should possess, and had an innate learning ability for interpreting molecular topology.

**Distinguishing of Conformer Rotations.** The ability to "know" the relative orientation of a molecule to its counterpart is also required for 3D DL models. This rotation-distinguishing ability is extremely important, if not required, in the biochemical analysis. For example, a real molecular-ligand finds its optimal relative-orientation to the binding partner (e.g., target protein) through an energy-minimization process (Figure 4a). Changes in the relative orientation alter the ligand affinity obviously, although the atom connectivities are all the same to the molecular rotations. It could be envisioned that a 3D DL model rotates a target protein and/or molecules in the 3D coordinate and decides which molecule is best candidate as a drug by estimating the binding affinity. It would be the first step towards this DL scenario to make a DL model predict the output differently with different ligand orientations for the prediction of the optimal pose.

We employed the *aligned* BACE dataset ($BACE_{algn}$) as our target for training the molecular orientations; the training could be achieved, because $BACE_{algn}$ provides the 3D coordinates of molecules aligned to the fixed position of the binding pocket of β-secretase 1, in addition to their experimental activeness. In this set of training, we adopted the 3D position of the conformer provided by the original report rather than generating it with MMFF94, and assessed the trained 3DGCN model with the rotated conformers from the test set. The training was conducted only with the aligned conformers; in other words, no rotated conformers were provided during training. It is also to note that additional information, such as the binding atoms in a molecule or the binding-pocket structure of β-secretase 1, was not provided during training.

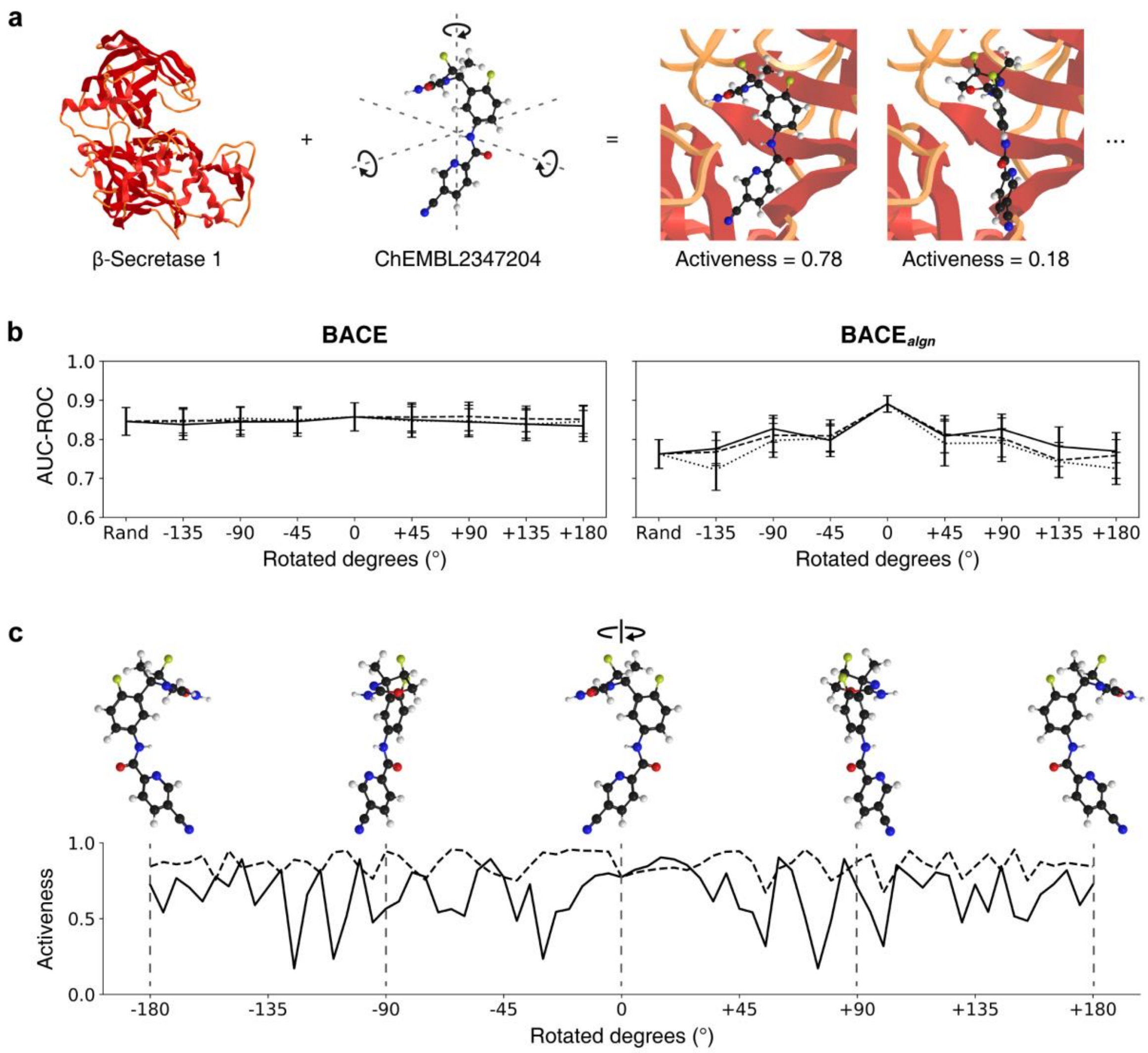

**Figure 4.** Activeness prediction with molecular poses. (A) Illustration of the conformer-rotation experiment for the BACE$_{algn}$ dataset. The left image shows the alignment of a molecule along the binding pocket of β-secretase 1, depicted with the crystal structure from the protein data bank 4J0V. The molecules are rotated gradually or randomly and provided to the previously trained model for activeness prediction, as depicted in the right image. Note that no rotated conformers are provided in the training. (B) Performance-evaluation results upon rotation of conformers from the test set on the BACE and BACE$_{algn}$ datasets. A trained model is provided with test molecules rotated along the x (solid), y (dashed), and z (dotted) axes, and the performance is measured. (C) Prediction results for ChEMBL2347204 molecule, is shared with both test sets, upon rotation on the z axis. Activeness fluctuation during rotation is observed in the model trained with the BACE$_{algn}$ dataset (solid), compared with the BACE dataset (dashed). Atom colors: black (carbon), white (hydrogen), red (oxygen), blue (nitrogen), and yellow (fluorine).

The test experiments showed that the 3DGCN training with BACE$_{algn}$ increased the AUC-ROC (from 0.857 to 0.891) and AUC-PR values (from 0.816 to 0.847), compared with the results with BACE, for the unrotated (i.e., 0° rotation) test sets. Of more significance was the results that the performance of the $BACE_{algn}$-trained 3DGCN model depended upon the relative orientation of a ligand, arguably confirming its orientation-distinguishing ability (Figure 4b, S4). The rotation degrees for a ligand changed the predicted values in activeness with greater changes in the larger rotation degrees (Table S3). These behaviors could be extrapolated from the real-world situation, where the more rotated, the more different in the local topology seen from the binding site of an enzyme; however, its realization in DL had not been achieved. The orientation-distinguishing ability of 3DGCN was investigated further with the molecules common to $BACE_{algn}$ and BACE. The results with a positive molecule (ChEMBL2347204) were presented as a representative in Figure 4c. The molecule was rotated from the provided orientation with 5° increment on the z-axis, and its activeness of inhibition against β-secretase 1 was predicted by the trained 3DGCN model, showing the high rotation-dependent fluctuations in the prediction. A negative molecule (ChEMBL1209418) also showed analogous fluctuation characteristics compared with the grain patterns from BACE (Figure S5). The fluctuations shown in the $BACE_{algn}$ training positively confirmed that the 3DGCN recognized the local 3D structures in the estimation of similarities to and/or differences from the trained patterns. In other words, the model did not predict the activeness only by looking at an entire molecule in the distance, and did understand the relative orientation of the molecule to the target protein.

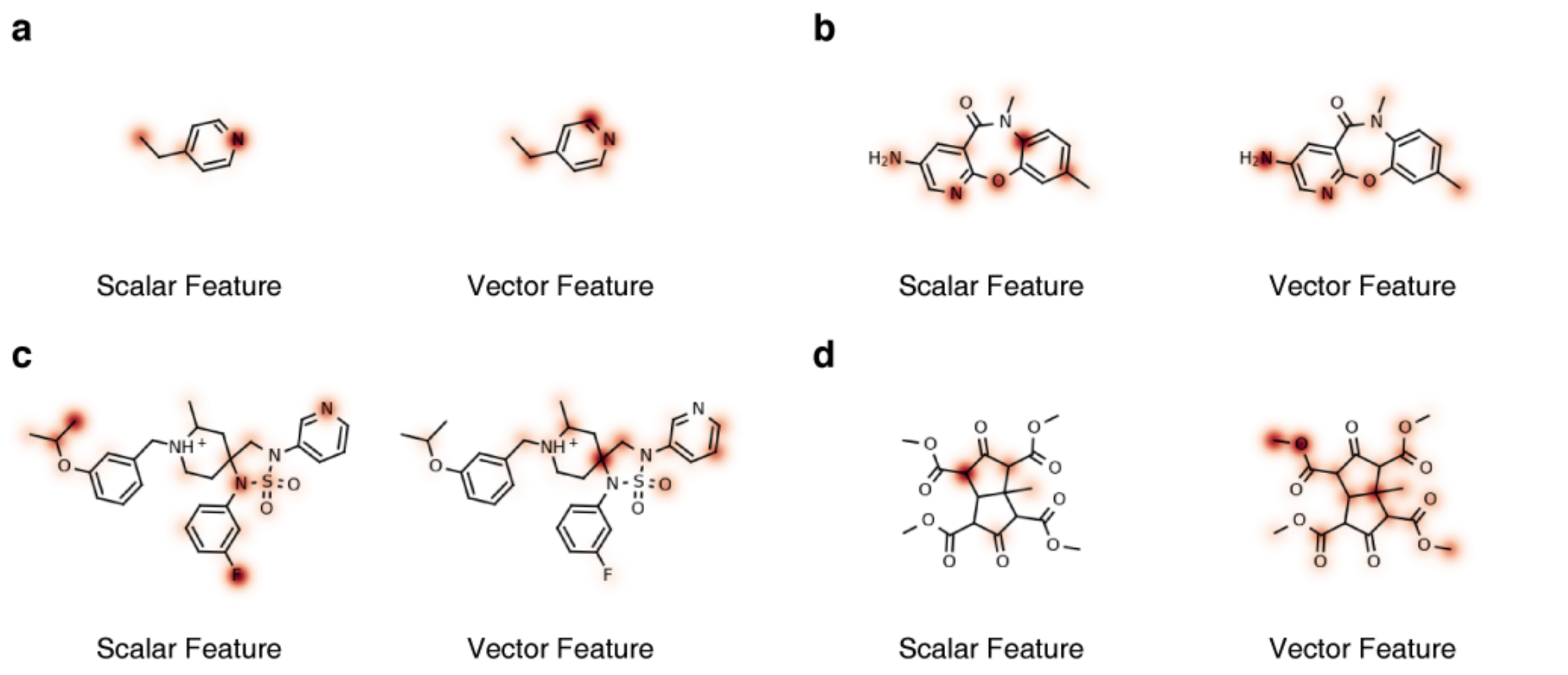

**Figure 5.** Contribution to molecular-feature generation. (A-D) Visualization of the influences of individual atom on molecular-feature generation after training for the FreeSolv (A), ESOL (B), BACE*algn* (C), and HIV (D) datasets. Atomic features are collected before aggregation process, and their relative ratio among atoms is depicted with the color intensity on the molecular structure. Example molecules are chosen randomly from the test set for each dataset.

**Atomic Contribution to Molecular Features.** We sought to investigate the inner workings of 3DGCN. As DL is innately the black box that hides its trained "brain" inside, we attempted to garner chemical insights from the black box by visualizing the generation process of molecular features with the atomic features that were distributed on a molecule. Specifically, a heat map for each molecule was made by coloring an atom in the molecule according to its contribution level for the molecular feature (Figure 5, S6-9). As the 3DGCN collected the scalar and vector features differently with the two molecular-feature generation strategies, heat maps were generated based on the used strategies. Interestingly, the contribution pattern of scalar features was very different from that of vector features for each atom, regardless of the datasets. The highlights in the contribution pattern were not only on functional groups from the carbon backbone, but also on diverse moieties including carbon chains and benzene rings. In addition, two functional groups of the same type in a molecule tended to have different contribution weights, even though they had the same connectivity in a range of second-nearest neighborhood. Taken together, these visualizations support the 3D-recognizing ability of 3DGCN.

## CONCLUSION

Molecular interactions take place in the 3D space and are highly influenced by the molecular conformation and relative orientation; however, little attention has yet been paid to the spatial information of molecules in the deep-learning (DL) chemistry. In this work, we demonstrated that the 3DGCN DL model, combining graph convolutional network (GCN) and 3D topology, accurately predicted the characteristics of molecules, local or nonlocal, from 3D molecular graphs. Our model, trained on four datasets in the chemical and biological fields, proved better in performance than the state-of-the-art DL models used in chemistry.

The significant advance made in this work is the generalizability and distinguishability of 3DGCN in the interpretation of 3D molecular conformation. Especially, its orientation distinguishability in the prediction of protein-ligand binding affinity would pave the way for the development of next-generation DL algorithms for 3D recognition, which has great impact on drug discovery and development. We believe that our findings provide a critical step towards facile DL applicability to problems in chemistry, as the first demonstration of the GCN that utilizes the full spatial topology of molecules on prediction. Many tasks still need to be investigated for full-fledged applications, including input feature selection and hyperparameter optimization, but this work suggests novel insights into molecular systems in DL chemistry, including spatial-structure formulation, aggregation strategy, and influence mapping. We also envision that the DL approach would act as a versatile (and complimentary) toolbox for tackling chemical problems.

## EXPERIMENTAL SECTION

**Datasets.** For our experiments, we employed regression and classification tasks from categories of chemical properties and biochemical interactions. Four datasets were chosen that are publicly available and commonly used: the Free Solvation Database (FreeSolv) is a dataset of 643 experimental and calculated hydration free energies optimized by general AMBER force field (GAFF);[34] the ESOL dataset is a collection of aqueous solubility in log(mol·$L^{-1}$) for 1128 compounds,[35] and targets the prediction of log-valued solubility from chemical structures provided as SMILES-encoded strings; the BACE dataset is a collection of experimental binary inhibition labels with the 3D conformer information for 1547 small-molecule inhibitors for human β-secretase 1, which is a potential therapeutic target for Alzheimer's disease;[44] the HIV dataset is a collection of 41127 molecules with the experimental inhibition results for replication of human immunodeficiency virus that were introduced by the Drug Therapeutics Program AIDS Antiviral Screen. The confirmed moderately active (CM) data were merged with the confirmed active (CA) data as active molecules, while the confirmed inactive (CI) data were used as inactive molecules.[45]

**3DGCN Architecture.** The 3DGCN consisted of three modules: convolutional layer, feature-aggregation layer, and fully connected layer. Upon initialization of the first level features, the scalar features of nodes were encoded as shown in Table S1, while the vector features were initialized with zeros. The first phase of the convolutional layer combined the two features from each node and generated intermediate features. In the second phase, the intermediate features were collected and summed along neighborhoods, leading to the generation of higher-level features. Through the two convolutional layers, the scalar and vector features were updated by neighborhood information, resulting in the information integration from the second nearest neighborhood. For activation, the ReLu function was used for all scalar-form outputs, and tanh was used for all vector-form outputs.

After convolution, the feature-aggregation layer collected the features along the nodes with two strategies, making node-independent, molecular features in order to provide permutational invariance for the model: the $3DGCN_{sum}$ summed all atom features distributed on the nodes, generating the molecular scalar and vector features; the $3DGCN_{max}$ selected the atom feature with the maximum value as a molecular feature. When finding the maximum with the vector features, the norm was used for the comparison.

The generated molecular features were fed to the fully connected neural networks with ReLu activation for prediction. The scalar feature was given to the two-layer stack of a fully connected neural network, while the vector feature was given to the two-layer stack of a time-distributed, fully connected neural network for the inhibition of separation between each axis during linear combinations. Finally, the outputs were flattened, concatenated, and fed into a single-layer neural network that predicted the experimental values from the datasets. The hyperparameters used in the overall 3DGCN models are shown in Table S3. No hyperparameter optimization was conducted except for the comparison between aggregation strategies.

**Model Training and Evaluation.** The 3DGCN was implemented in Python using the open-source machine learning library Keras[46] 2.1.6 with TensorFlow[47] 1.5 as a backend. Training was controlled by learning-rate decay and early-stopping techniques, which observed the validation error to lower the learning rate or stop the training. The learning rate was decreased by a factor of 0.9 when the loss reached a plateau, with a patience of 10, and the termination was determined with a patience of 15. Enough epochs were set to prevent termination by the end of epochs, not through the early-stopping mechanism. Models were evaluated by RMSE or AUC-ROC with 10-fold stratified cross-validation. For each fold, the dataset was randomly split into training set (80 %), validation set (10 %), and test set (10 %). All 3DGCN models were trained and executed on NVIDIA GTX 1080Ti GPU.

**Evaluation of Rotation Effects.** Given trained models from the cross-validation, molecules from their test sets were randomly or gradually rotated along the x, y, and z axes and saved as alternative test sets. The random rotation was conducted on all the axes together at once (e.g., 34° on x, 10° on y, and 143° on z), while the stepwise rotation was done independently with 45° increment. The alternative test sets were then evaluated by the model without additional training process and averaged. For single-molecule rotation test, a molecule was randomly selected from the test set and rotated at 5° increment along the z axis, and the output was predicted with the trained model.

**Visualization of Atom Influence.** To calculate the influence of individual atoms on a molecular feature, the outputs before aggregation process were taken from the trained model. For calculating the weights of influence of the atoms in a molecule in the $3DGCN_{sum}$, the scalar or vector feature of each and every atom was divided by the summed scalar or vector features across the entire molecule, respectively. For the vector-feature calculation, the norm was used. On the other hand, in the $3DGCN_{max}$, the weight of influence for an atom was calculated by dividing its feature occurrence in the molecular features by the total number of molecular features. The calculated weights of influence were used for visualizing the heat map of a molecule.

## AUTHOR INFORMATION

**Corresponding Author**

*ischoi@kaist.ac.kr

**ORCID**

Insung S. Choi: 0000-0002-9546-673X

**Author Contributions**

H.C. conceived and designed the 3DGCN and experiments. H.C. and I.S.C. performed the analyses and wrote the paper. All authors discussed the results and commented on the manuscript.

**Notes**

The authors declare no competing financial interest.

## ACKNOWLEDGMENTS

H.C. thanks Jingun Jung for helpful discussion on idea inception. This work was supported by the Basic Science Research Program through the National Research Foundation of Korea (NRF) funded by the Ministry of Science, ICT & Future Planning (MSIP 2012R1A3A2026403) and K-Valley RED&B Program of KAIST.

**SUPPORTING INFORMATION**

# Three-Dimensionally Embedded Graph Convolutional Network (3DGCN) for Molecule Interpretation

Hyeoncheol Choi and Insung S. Choi*
Department of Chemistry, KAIST, Daejeon 34141, Korea

**Table S1. Atom features in the initial representation of molecules.**

| Feature | Description | Type | Size |
|---|---|---|---|
| Atom type | atom type | one-hot | 15 |
| Degree | the number of heavy atom neighbors (0 to 6) | one-hot | 7 |
| Number of hydrogens | the number of neighboring hydrogens (0 to 4) | one-hot | 5 |
| Implicit valence | the number of implicit hydrogens (0 to 6) | one-hot | 7 |
| Hybridization | $sp$, $sp^2$, $sp^3$, $sp^3d$, or $sp^3d^2$. | one-hot | 5 |
| Formal charge | atomic formal charge (-3 to +3) | one-hot | 7 |
| Ring size | whether this atom belongs to a ring (ring size: 3 to 8) | binary | 6 |
| Aromaticity | whether this atom is part of an aromatic system. | binary | 1 |
| Chirality | *R*, *S*, or nonchiral | one-hot | 3 |
| Acid/base | whether this atom is acidic or basic | binary | 2 |
| Hydrogen bonding | whether this atom is a hydrogen bond donor or acceptor | binary | 2 |
| Total | | | 60 |

**Table S2. Evaluation results upon rotation of the test set for the BACE and BACE$_{algn}$ datasets. All results are obtained through ten-fold cross-validation.**

| Axis | Degree | BACE | | BACE$_{algn}$ | |
|---|---|---|---|---|---|
| | | AUC-ROC | AUC-PR | AUC-ROC | AUC-PR |
| | 0° | 0.857 ± 0.036 | 0.816 ± 0.037 | 0.891 ± 0.021 | 0.847 ± 0.024 |
| x | 45° | 0.850 ± 0.033 | 0.803 ± 0.029 | 0.808 ± 0.043 | 0.752 ± 0.059 |
| | 90° | 0.844 ± 0.034 | 0.800 ± 0.035 | 0.826 ± 0.029 | 0.761 ± 0.046 |
| | 135° | 0.839 ± 0.036 | 0.780 ± 0.046 | 0.781 ± 0.051 | 0.718 ± 0.074 |
| | 180° | 0.834 ± 0.040 | 0.789 ± 0.047 | 0.770 ± 0.030 | 0.710 ± 0.057 |
| | 225° | 0.838 ± 0.039 | 0.779 ± 0.044 | 0.776 ± 0.043 | 0.710 ± 0.051 |
| | 270° | 0.845 ± 0.037 | 0.797 ± 0.042 | 0.827 ± 0.034 | 0.774 ± 0.038 |
| | 315° | 0.845 ± 0.038 | 0.790 ± 0.037 | 0.797 ± 0.042 | 0.749 ± 0.051 |
| y | 45° | 0.857 ± 0.036 | 0.818 ± 0.033 | 0.813 ± 0.048 | 0.738 ± 0.066 |
| | 90° | 0.858 ± 0.037 | 0.819 ± 0.030 | 0.826 ± 0.029 | 0.761 ± 0.046 |
| | 135° | 0.852 ± 0.033 | 0.812 ± 0.030 | 0.747 ± 0.045 | 0.675 ± 0.076 |
| | 180° | 0.851 ± 0.036 | 0.808 ± 0.032 | 0.758 ± 0.059 | 0.694 ± 0.089 |
| | 225° | 0.848 ± 0.032 | 0.796 ± 0.036 | 0.767 ± 0.030 | 0.698 ± 0.051 |
| | 270° | 0.847 ± 0.035 | 0.796 ± 0.027 | 0.810 ± 0.043 | 0.745 ± 0.056 |
| | 315° | 0.847 ± 0.032 | 0.800 ± 0.034 | 0.809 ± 0.041 | 0.745 ± 0.048 |
| z | 45° | 0.847 ± 0.042 | 0.807 ± 0.042 | 0.789 ± 0.058 | 0.725 ± 0.075 |
| | 90° | 0.847 ± 0.040 | 0.800 ± 0.041 | 0.791 ± 0.036 | 0.730 ± 0.051 |
| | 135° | 0.838 ± 0.041 | 0.800 ± 0.043 | 0.742 ± 0.041 | 0.680 ± 0.070 |
| | 180° | 0.846 ± 0.039 | 0.803 ± 0.046 | 0.725 ± 0.041 | 0.638 ± 0.064 |
| | 225° | 0.845 ± 0.036 | 0.801 ± 0.051 | 0.722 ± 0.053 | 0.639 ± 0.084 |
| | 270° | 0.854 ± 0.030 | 0.802 ± 0.038 | 0.797 ± 0.043 | 0.738 ± 0.062 |
| | 315° | 0.850 ± 0.034 | 0.803 ± 0.042 | 0.802 ± 0.033 | 0.750 ± 0.044 |
| x, y, z | random | 0.846 ± 0.036 | 0.795 ± 0.037 | 0.762 ± 0.037 | 0.690 ± 0.059 |

**Table S3. Hyperparameters used for 3DGCN training.**

| Group | Hyperparameter | Size |
|---|---|---|
| 3DGCN | number of output units for scalar-to-scalar operation | 128 |
| | number of output units for vector-to-scalar operation | 128 |
| | number of output units for scalar-to-vector operation | 128 |
| | number of output units for vector-to-vector operation | 128 |
| | number of output units for scalar convolution | 128 |
| | number of output units for vector convolution | 128 |
| | number of convolution layers | 2 |
| Fully connected | number of output units on fully connected layers | 128 |
| | number of fully connected layers | 2 |
| Training | batch size | 8 or 16 |
| | initial learning rate | 0.001 |
| | minimum learning rate | 0.0005 |
| | gradient descent method | adam |

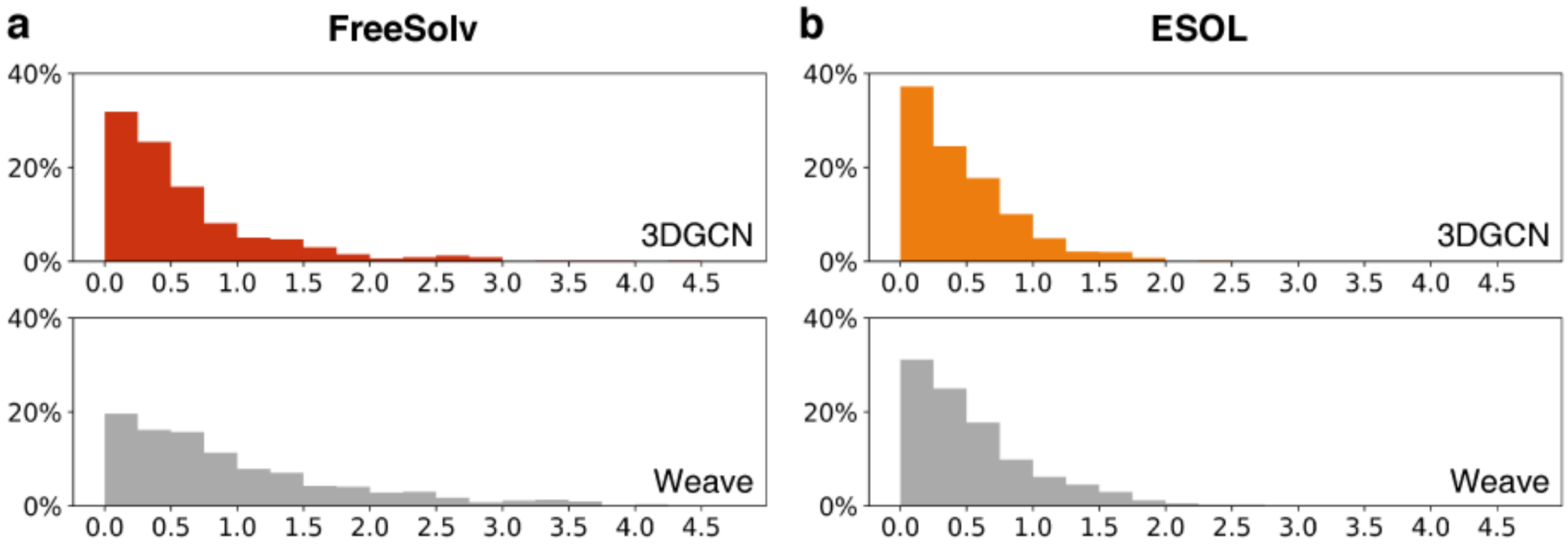

**Figure S1.** Error distributions for FreeSolv and ESOL. (A-B) Histograms of absolute errors from the test sets for the FreeSolv (A) and ESOL (B) datasets of 3DGCN and Weave models.

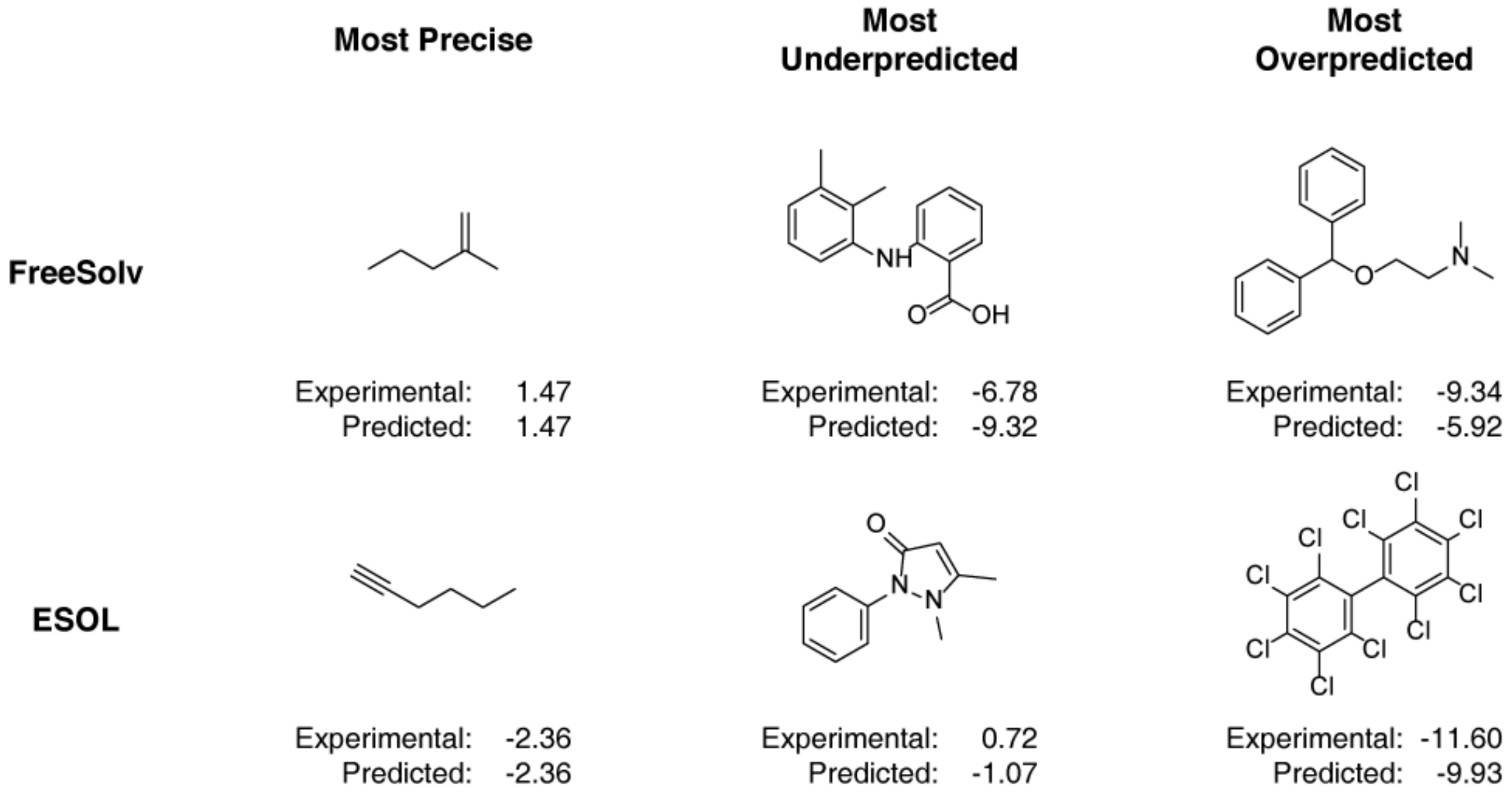

**Figure S2.** Example molecules and corresponding prediction results from the FreeSolv and ESOL datasets. Most precise, underpredicted, and overpredicted molecules are chosen from the test set.

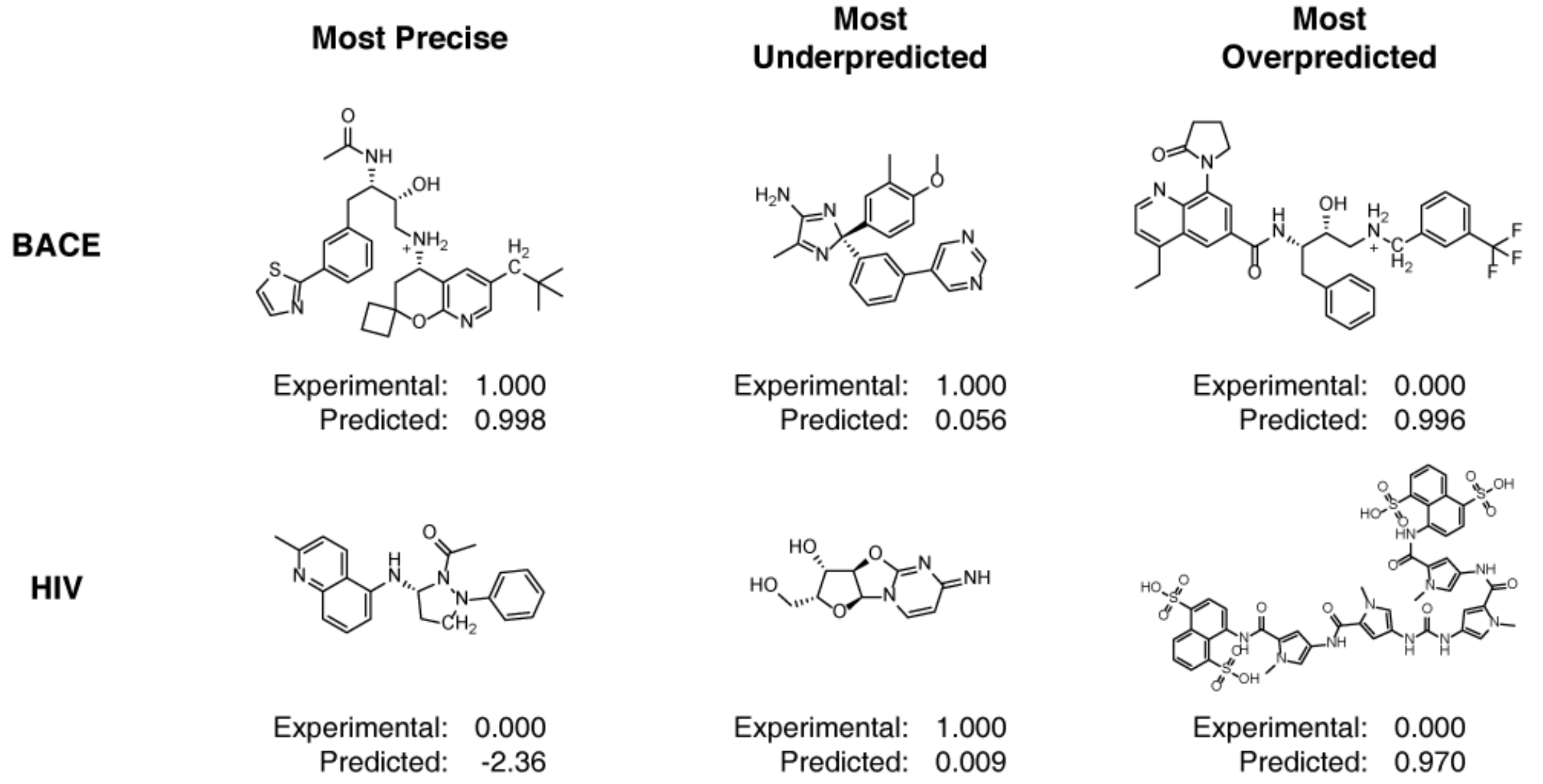

**Figure S3.** Example molecules and corresponding prediction results from the BACE and HIV datasets. Most precise, underpredicted, and overpredicted molecules are chosen from the test set.

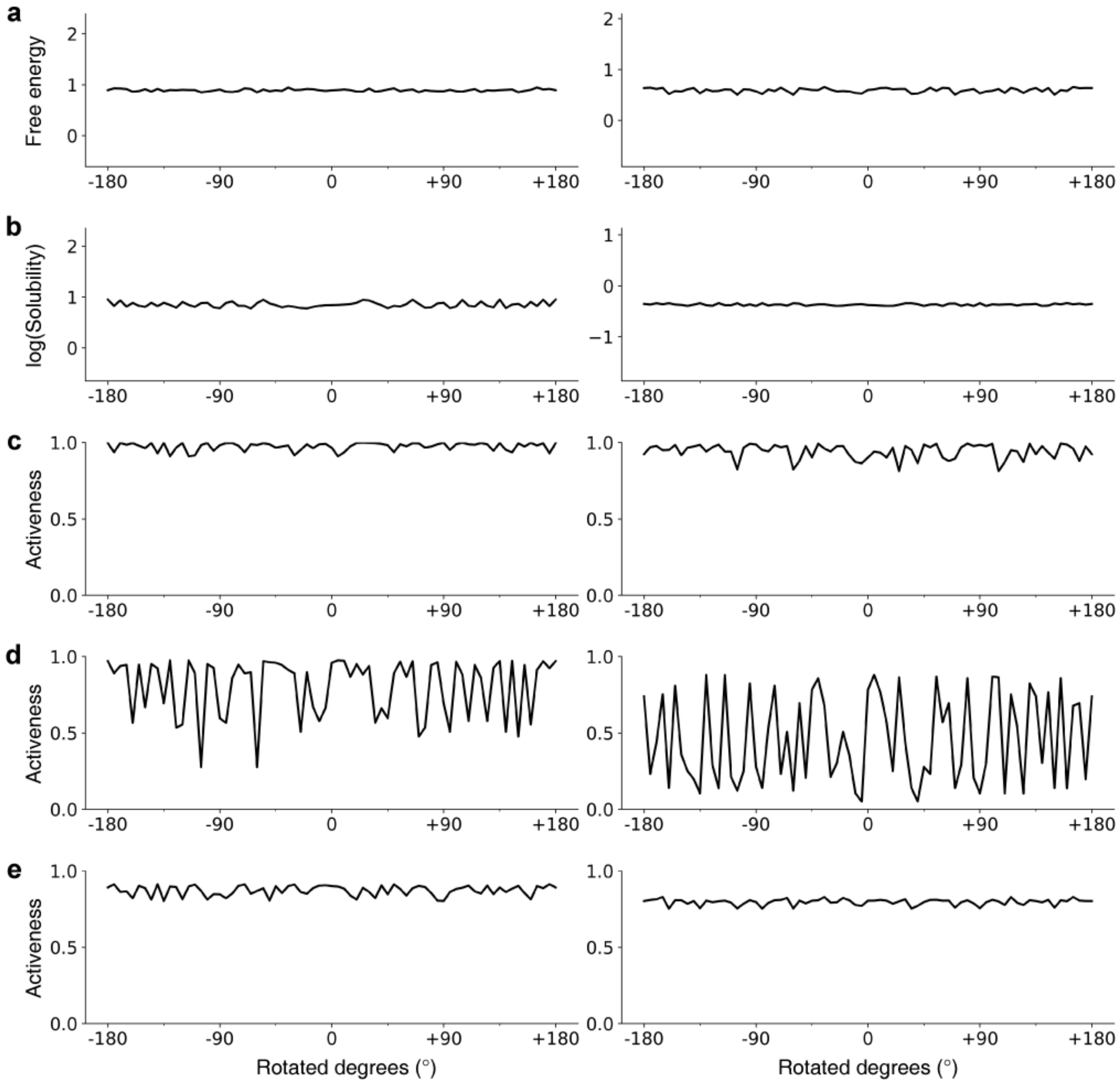

**Figure S4.** Rotational equivariance. (A-E) Prediction results for two molecules, randomly selected from the test sets of the FreeSolv (A), ESOL (B), BACE (C), BACE$_{algn}$ (D), and HIV (E) datasets, upon rotation on the z axis.

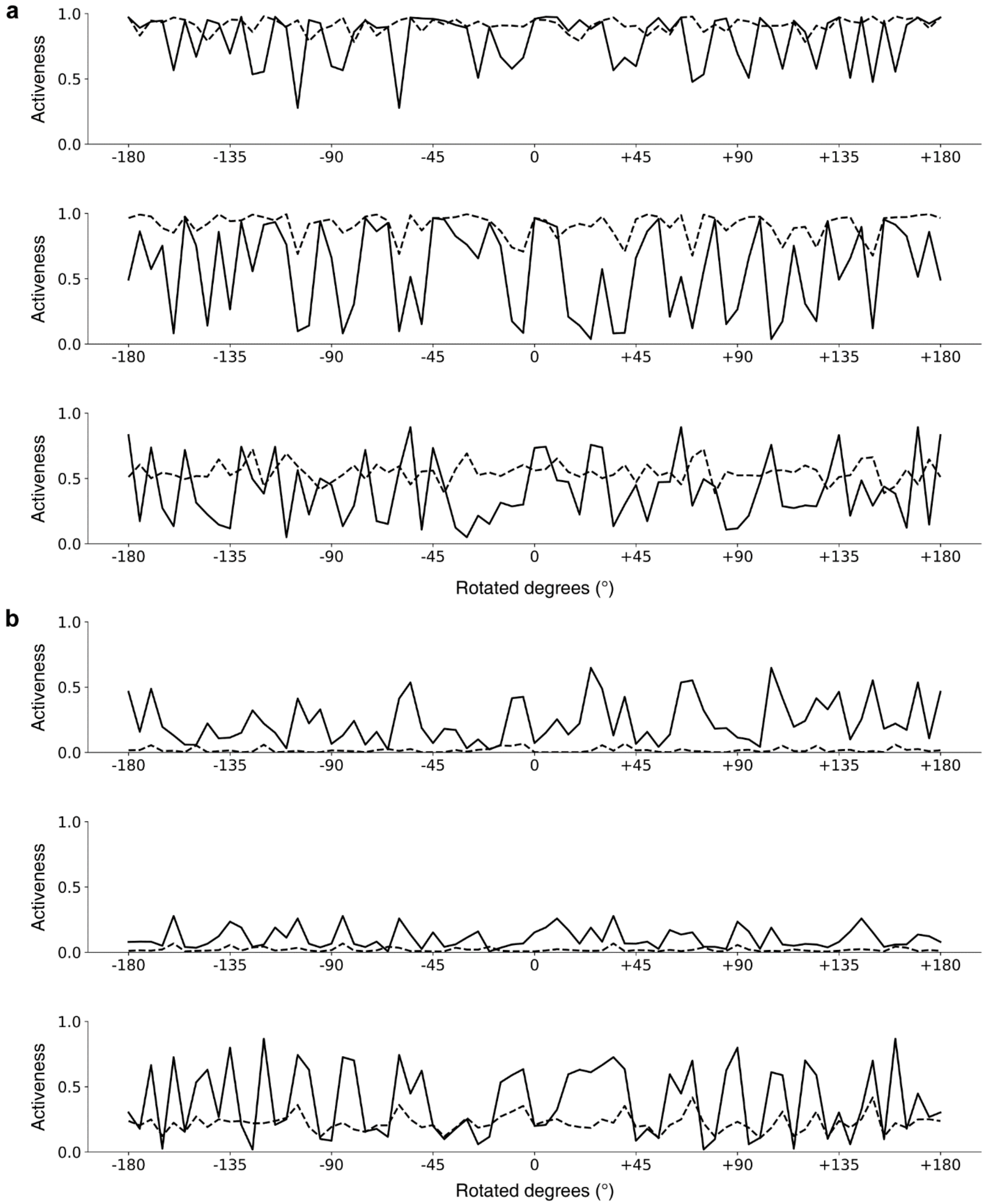

**Figure S5.** Comparison between BACE and BACE$_{algn}$ for rotation-activeness prediction. (A-B) Prediction vs. rotated degree for molecules of the BACE$_{algn}$ (solid) and BACE (dashed) datasets. Positive (A) and negative (B) molecules are selected randomly from the test set.

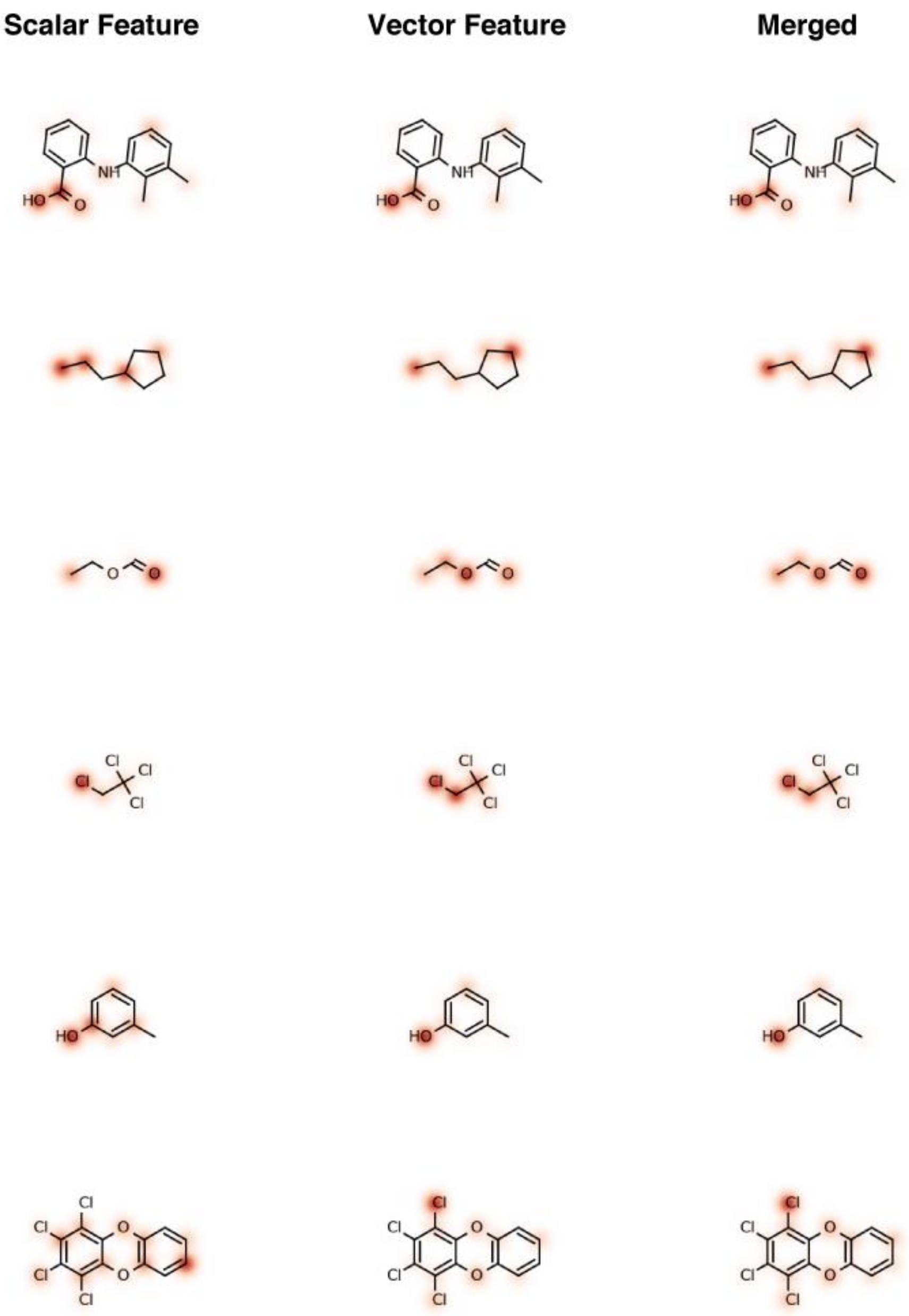

**Figure S6.** Visualization of influence of individual atom on molecular-feature generation after training with FreeSolv.

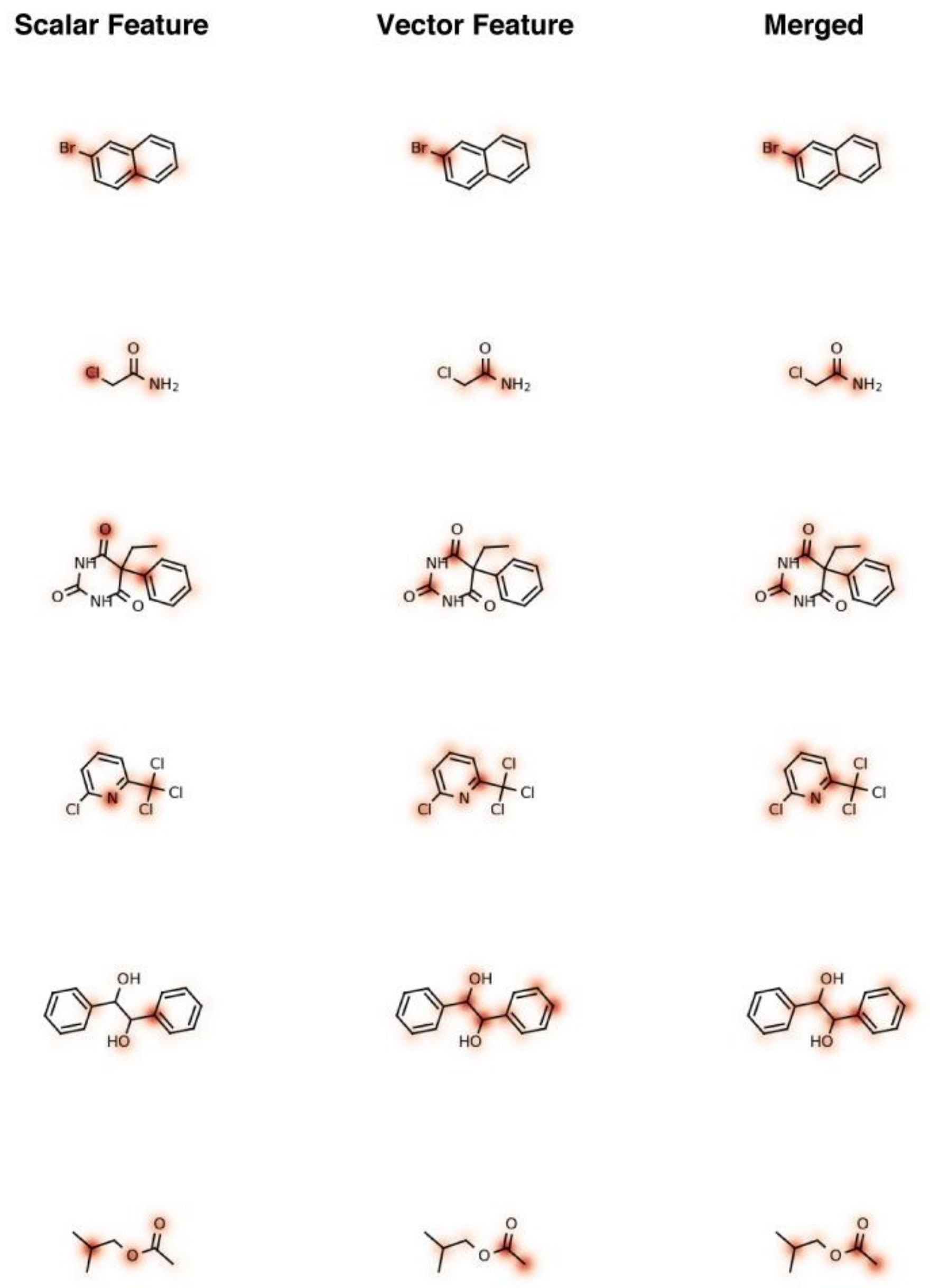

**Figure S7.** Visualization of influence of individual atom on molecular-feature generation after training with ESOL.

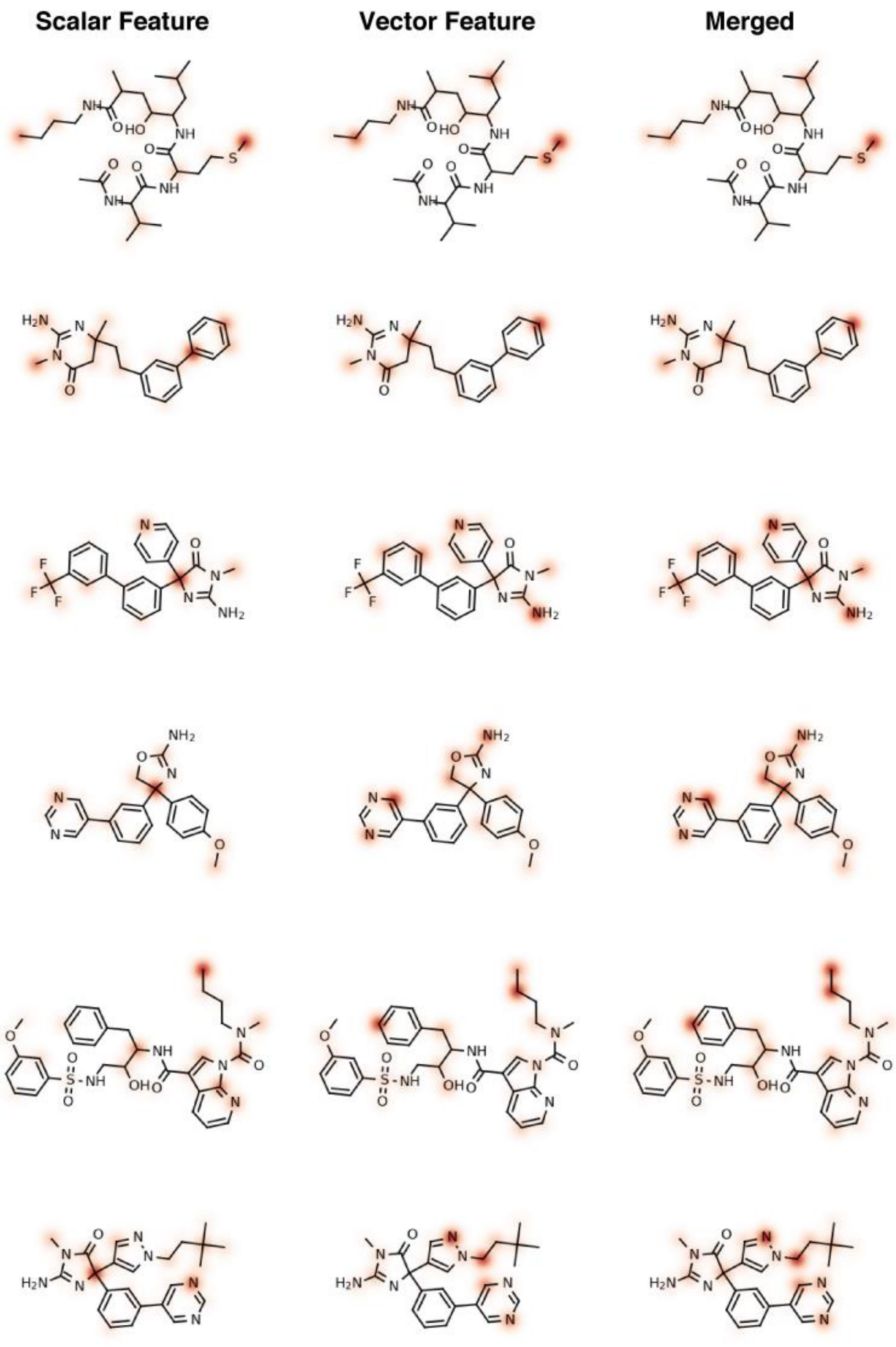

**Figure S8.** Visualization of influence of individual atom on molecular-feature generation after training with BACE.

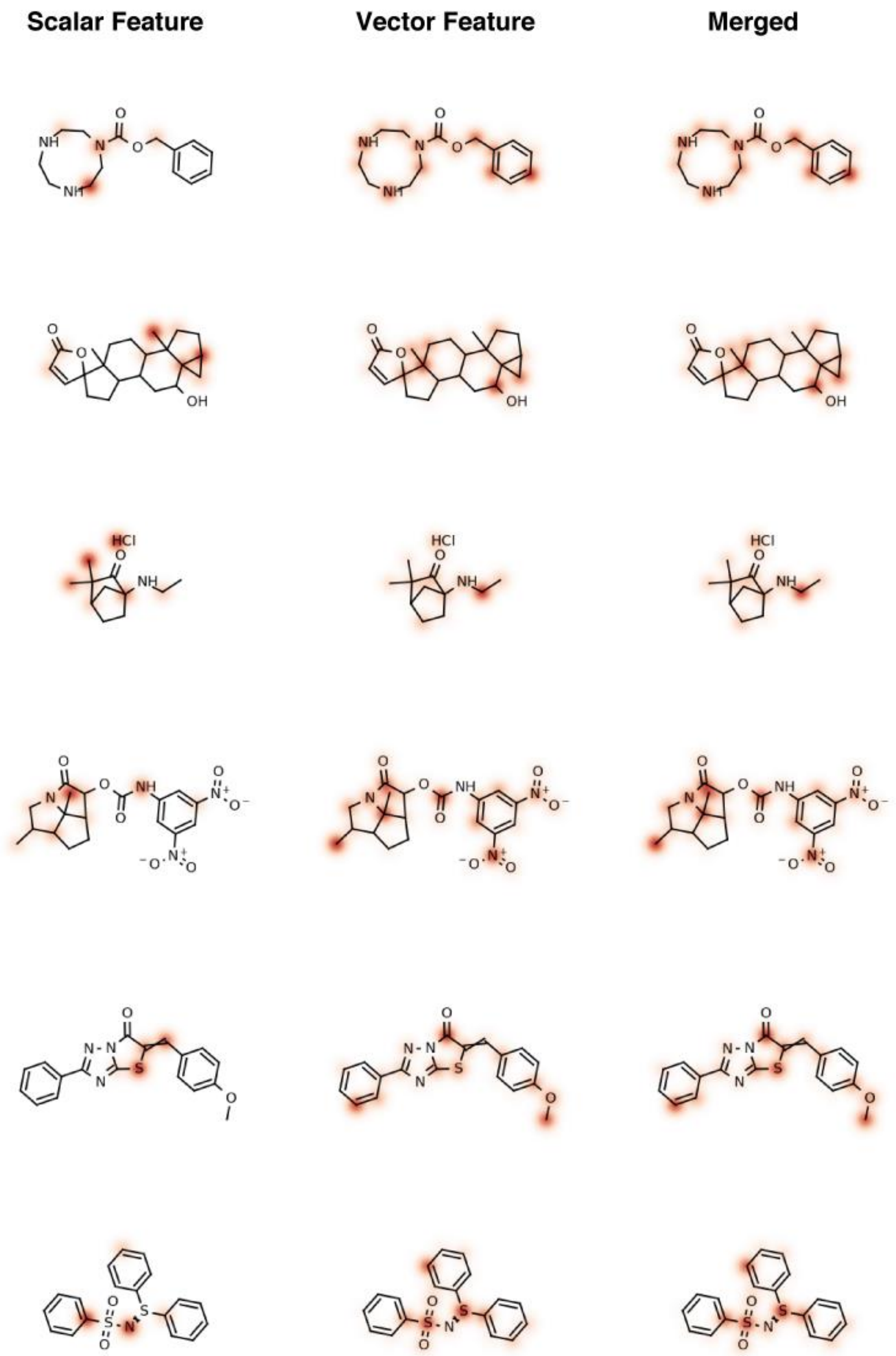

**Figure S9.** Visualization of influence of individual atom on molecular-feature generation after training with HIV.